\documentclass[10pt,journal,compsoc]{IEEEtran}
\ifCLASSOPTIONcompsoc
  \usepackage[nocompress]{cite}
\else
  \usepackage{cite}
\fi
\usepackage{url}
\usepackage{graphicx}
\usepackage{booktabs}
\usepackage{multirow}
\usepackage{stfloats}
\usepackage{makecell}
\usepackage{bbding} 
\usepackage{colortbl}
\usepackage{picins}
\usepackage{color,xcolor}
\usepackage{soul}

\sethlcolor{yellow}
\usepackage{amsmath}
\usepackage{amssymb}
\definecolor{mygray}{gray}{0.8} 
\soulregister\cite7
\soulregister\ref7
\ifCLASSINFOpdf
\else
\fi
\begin{document}
%
\title{MagicSeg: Open-World Segmentation Pretraining via Counterfactural Diffusion-Based Auto-Generation}

\author{Kaixin Cai*, Pengzhen Ren*, Jianhua Han, Yi Zhu, Hang Xu, \\Jianzhuang Liu, Xiaodan Liang$^{\dag}$

\IEEEcompsocitemizethanks{
\IEEEcompsocthanksitem {*} denotes equal contribution.
\IEEEcompsocthanksitem {\dag} Xiaodan Liang is the corresponding author. 
\IEEEcompsocthanksitem Kaixin Cai is with the Shenzhen Campus of
Sun Yat-sen University, Shenzhen, China. 
E-mail:caikx7@mail2.sysu.edu.cn.
\IEEEcompsocthanksitem Pengzhen Ren is with PengCheng Laboratory.
E-mail: pzhren@foxmail.com.
\IEEEcompsocthanksitem Jianhua Han, Yi Zhu, Hang Xu, and Jianzhuang Liu are with Huawei Noah’s Ark Lab.
E-mail: {hanjianhua4, zhuyi36, xu.hang}@huawei.com, jz.liu@siat.ac.cn.
\IEEEcompsocthanksitem  Xiaodan Liang is with the Shenzhen Campus of Sun Yat-sen University,
Shenzhen, China, and also with PengCheng Laboratory.
E-mail: xdliang328@gmail.com.
}

}
%
%

\markboth{Journal of \LaTeX\ Class Files,~Vol.~14, No.~8, August~2015}%
{Shell \MakeLowercase{\textit{et al.}}: Bare Demo of IEEEtran.cls for Computer Society Journals}
%



\IEEEtitleabstractindextext{%
\begin{abstract}
Open-world semantic segmentation presently relies significantly on extensive image-text pair datasets, which often suffer from a lack of fine-grained pixel annotations on sufficient categories. The acquisition of such data is rendered economically prohibitive due to the substantial investments of both human labor and time.
In light of the formidable image generation capabilities of diffusion models, we introduce a novel diffusion model-driven pipeline for automatically generating datasets tailored to the needs of open-world semantic segmentation, named “\textbf{MagicSeg}”. Our MagicSeg initiates from class labels and proceeds to generate high-fidelity textual descriptions, which in turn serve as guidance for the diffusion model to generate images. Rather than only generating positive samples for each label, our process encompasses the simultaneous generation of corresponding negative images, designed to serve as paired counterfactual samples for contrastive training. Then, to provide a self-supervised signal for open-world segmentation pretraining, our MagicSeg integrates an open-vocabulary detection model and an interactive segmentation model to extract object masks as precise segmentation labels from images based on the provided category labels. By applying our dataset to the contrastive language-image pretraining model with the pseudo mask supervision and the auxiliary counterfactual contrastive training, the downstream model obtains strong performance on open-world semantic segmentation.  We evaluate our model on PASCAL VOC, PASCAL Context, and COCO, achieving SOTA with performance of 62.9\%, 26.7\%, and 40.2\%, respectively, demonstrating our dataset's effectiveness in enhancing open-world semantic segmentation capabilities. Project website: \url{https://github.com/ckxhp/magicseg}.
\end{abstract}    

\begin{IEEEkeywords}
Open-World Semantic Segmentation, Dataset synthesis, Counterfactual
\end{IEEEkeywords}}

\maketitle

\IEEEdisplaynontitleabstractindextext

%
\IEEEpeerreviewmaketitle

\IEEEraisesectionheading{\section{Introduction}\label{sec:introduction}}
\IEEEPARstart{T}{he} advent of vision-language models has given rise to open-world segmentation models, which exploit the correlation between image and text for segmentation purposes. These models derive their segmentation capabilities from vast reservoirs of image-text pairs. However, owing to the lack of pixel-level annotated data, their segmentation ability is comparatively circumscribed in contrast to closed-set semantic segmentation paradigms. To enhance the segmentation capabilities of open-world segmentation models, a dataset with a diverse set of categories and pixel-level annotations is essential,  while creating such a dataset demands considerable investment in terms of human resources. 

\begin{figure}[t]
    \centering
    \includegraphics[width=1.0 \linewidth]{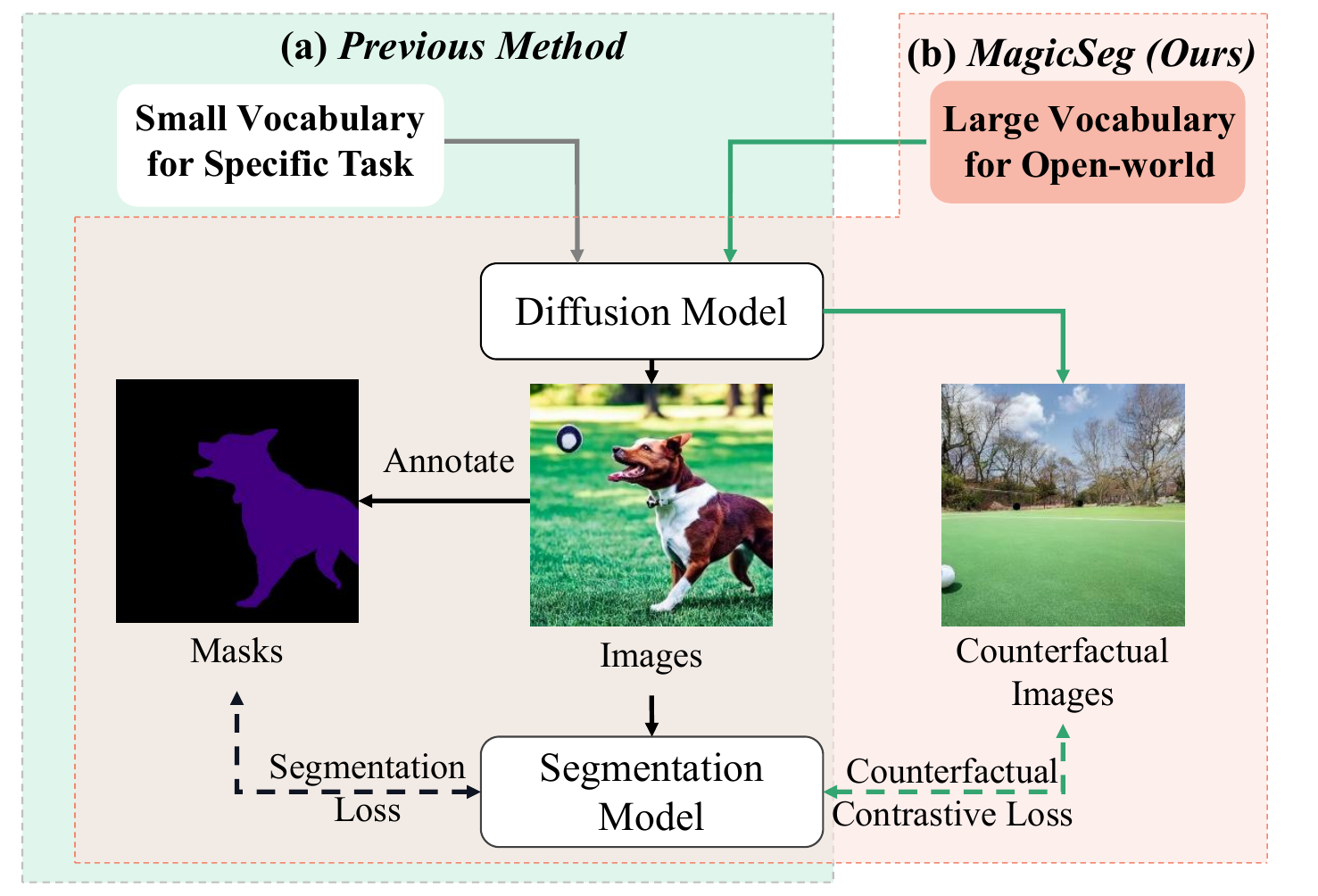}
    \caption{{Comparison between the previous method and MagicSeg. Previous methods \cite{DiffuMask, datediff} involved constructing dedicated segmentation datasets for specific downstream tasks. In contrast, MagicSeg leverages a large-scale vocabulary to build a dataset applied in open-world segmentation. Additionally, we generate counterfactual images to assist for segmentation task. }}
    \label{fig:intro_comp}
\end{figure}
To address this challenge, a viable approach is to leverage synthetic data and automatically obtain pixel-level annotations. Some research focuses on using Generative Adversarial Networks (GANs) \cite{GAN} to generate segmentation data, such as DatasetGAN \cite{dataetgan} and BigDatasetGAN \cite{bigdatasetgan}. However, they still require a small portion of real data with segmentation labels. In recent years, Diffusion models \cite{ddpm} have demonstrated remarkable capabilities in text-to-image generation tasks, as exemplified by models like DALL-E2 \cite{dalle} and Stable Diffusion \cite{stable}. Therefore, recent efforts aim to acquire semantic segmentation datasets using diffusion models. DatasetDM \cite{DATASETDM} employs an additional decoder trained on a small part of data to annotate generated data, while FreeMask \cite{freemask} enhances training by generating images from existing pixel labels. Both of these approaches still rely on the presence of real data with labels. Diffumask \cite{DiffuMask} and Dataset Diffusion \cite{datediff} utilize attention modules within diffusion models to construct pixel labels. However, labels generated from attention mechanisms may have relatively significant noise. Moreover, these studies primarily focus on constructing datasets for specific closed sets like PASCAL VOC  (Figure \ref{fig:intro_comp} (a)), without endeavors to create segmentation datasets suitable for open-world scenarios.

In this paper, we present MagicSeg, a novel yet streamlined pipeline for the construction of datasets tailored to open-world segmentation. We deploy this pipeline in the training of open-world segmentation models. Central to the dataset construction process is the imperative of curating a sufficiently diverse set of categories and images. To achieve this, we collect an extensive vocabulary of category names (Figure \ref{fig:intro_comp} (b)), 
 and leverage large language models such as ChatGPT to produce high-quality image descriptions for class names from our vocabulary through a series of carefully defined text generation conditions.

\begin{figure*}[t]
    \centering
    \includegraphics[width=1.0 \linewidth]{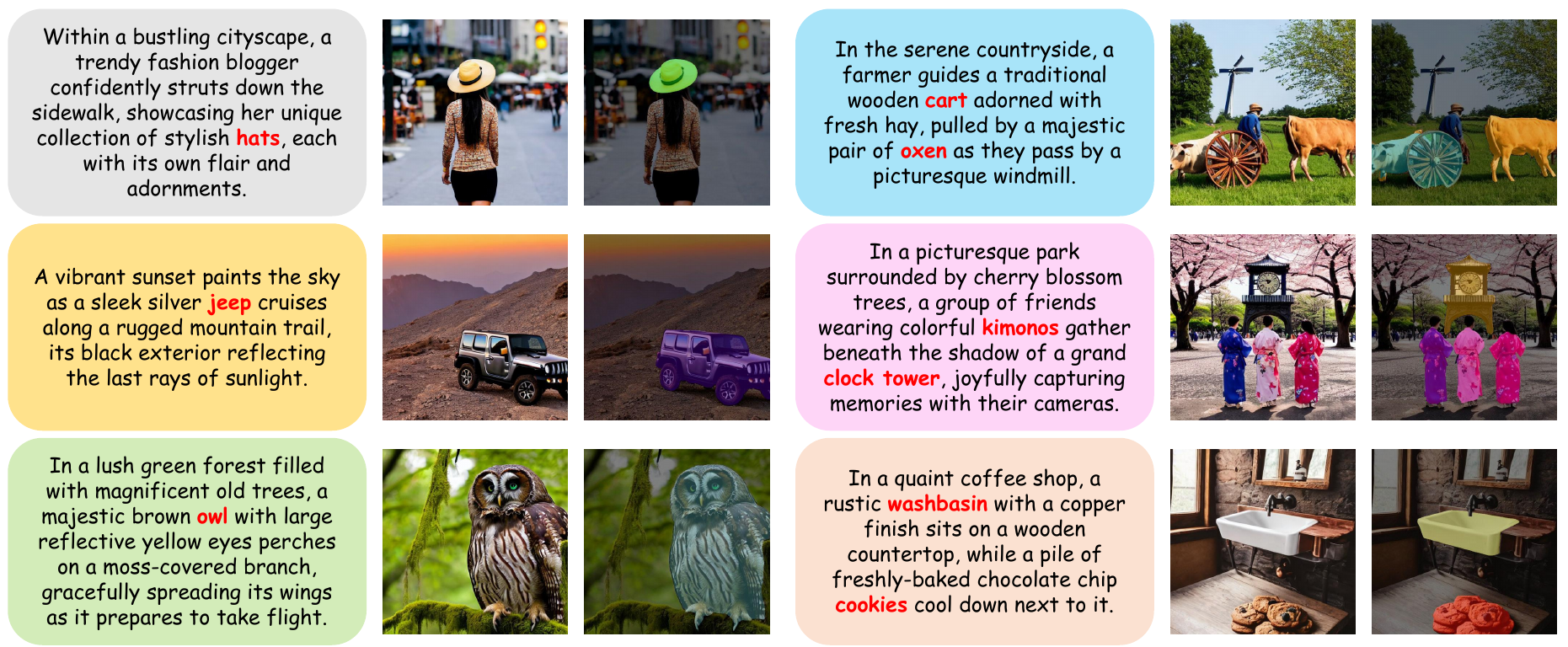}
    \caption{Visualization of synthetic image and mask from MagicSeg. With ChatGPT, we generate richly descriptive text from class names, which are used to create diverse images and corresponding masks.}
    \label{fig:intro}
\end{figure*}


The generation of high-quality pixel-level annotations is a challenge due to the high requirements for the fine-grained recognition ability of models. Therefore, given that the generated text contains class names, instead of using the attention mechanism of the diffusion model, which is not designed for segmentation, to annotate, we employ an open-vocabulary detector (\emph{i.e.} Grounding DINO \cite{liu2023grounding}) and an interactive segmentation model (\emph{i.e.} SAM \cite{kirillov2023segany}) to derive masks corresponding to target objects in the images. The generated dataset can be seen in Figure \ref{fig:intro}.
{This advantage is not available when using real images, as they lack information about known image categories. Compared to using real images for category labeling and then generating masks, synthetic data allows for controllable categories. Moreover, synthetic data can supplement diversity-rich images for rare categories in the real world, making it suitable for large-scale automated dataset construction and thus serving as a supplement to real datasets.}

During the training phase, three principal challenges emerge: (1) Noise in pixel-level annotations, such as missing object masks in the images; (2) A diversity of categories within the vocabulary, yet a limitation on the number of categories per image, results in a imbalance in the gradient between the existing and non-existent categories for each image; (3) Additionally, we contend with the imperative to mitigate significant biases towards categories in the vocabulary, which could affect the open-world segmentation capabilities of downstream models.

To address the first challenge, we introduce a counterfactual image construction method, generating images devoid of the target category in contrast to the original images for counterfactual contrastive training (Figure \ref{fig:intro_comp} (b)). With the self-supervised learning between counterfactual and original image, the model can develop object localization capabilities without reliance on masks which we annotate. 

For the second challenge, we propose a category random sampling strategy. During training, categories are randomly sampled from vocabulary, ensuring that the model employs a different subset of categories in each training iteration to predict masks, thereby relieving the gradient imbalance. In addition, this strategy effectively alleviates computational burdens as we do not need to predict masks with the full vocabulary for each image during the training phase.

For the third challenge, counterfactual contrastive training and category random sampling strategy can work together to solve it. With counterfactual contrastive training, the model can obtain class-agnostic segmentation ability because of the self-training for counterfactual image pair. Meanwhile, category random sampling strategy trains the model with different subsets each time, alleviating the overfitting problem of the model for the vocabulary categories.

In general, MagicSeg proposes a framework for generating open-world segmentation datasets, effectively narrowing the gap between open-world training and supervised training. To summarize, our contributions are as follow:
\begin{itemize}
    \item We present MagicSeg, a method designed to construct datasets for open-world segmentation. This dataset not only features a wide variety of categories, including pixel-level masks but also incorporates text descriptions and counterfactual images for noise mitigation.

    \item For the constructed dataset, we propose the category random sampling strategy and counterfactual contrastive training. These methods effectively enable the model to learn open-world segmentation capabilities.

    \item Experimental results demonstrate the efficacy of MagicSeg, achieving remarkable performance across multiple prominent datasets PASCAL VOC, PASCAL Context, COCO, and LVIS.
\end{itemize}
~\\
~\\
\section{Related Work}
\textbf{Open-world Segmentation.} The open-world problem primarily aims to get models with the ability to recognize new concepts through training on a closed-world dataset. Similar settings have been studied in recognition tasks \cite{bendale2015towards, liu2019large}, object detection \cite{wang2021unidentified, joseph2021towards}, and semantic segmentation \cite{ICLIP, xu2022groupvit, viewco, nakajima2019incremental, wysoczanska2024clip, Mixreorg, karazija2023diffusion, cha2023learning}. For segmentation, it necessitates a fine-grained recognition capability for new concepts, involving the pixel-wise segment ability of target objects. \cite{nakajima2019incremental} employs clustering based on pixel perception to achieve open-world unsupervised segmentation. More recently, with the development of large-scale pre-trained language-image models, such as CLIP \cite{CLIP}, which achieves robust zero-shot recognition by training on a large scale of image-text pairs, there have been efforts to use fine-grained semantic alignment from extensive image-text data. CLIP-DIY \cite{wysoczanska2024clip} applies the CLIP classification abilities on multi-scale of image to achieve segmentation. OVDiff \cite{karazija2023diffusion} uses the generative model to extract prototypes for segmentation. GroupViT \cite{xu2022groupvit} introduces learnable group tokens to achieve alignment between segments and text, while Viewco \cite{viewco} combines contrastive training across multiple views of images to enhance the model's segmentation capabilities. {MaskCLIP\cite{zhou2021denseclip} leverages CLIP features for annotation-free and zero-shot semantic segmentation. GEM \cite{gem} proposes a module which enables zero-shot open-vocabulary object localization using pre-trained vision-language models through self-self attention. CLIP Surgery \cite{li2023clipsur} improves the explainability of CLIP models by addressing opposite visualization and noisy activations. SCLIP \cite{wang2024sclip} enhances CLIP's performance for semantic segmentation by introducing a novel Correlative Self-Attention (CSA) mechanism. Nevertheless, the lack of pixel-level annotations in image-text data results in limited segmentation performance. Building based on image-text pre-trained models, our approach constructs a reliable synthetic segmentation dataset to enhance the model's open-domain segmentation capabilities.}

\noindent\textbf{Text-to-Image Synthesis.} There are several types of image generation models, including Variational AutoEncoder (VAE) \cite{vae}, flow-based models \cite{flow}, Generative Adversarial Networks (GANs) \cite{GAN} and diffusion models \cite{ddpm}. Recently, diffusion models have garnered significant attention due to their remarkable performance in text-to-image generation tasks. There are currently many outstanding diffusion-based image generation models, such as GLIDE \cite{nichol2021glide}, DALL-E2 \cite{dalle}, Imagen \cite{imagen}, and Stable Diffusion \cite{stable}. These models \cite{nichol2021glide, dalle, imagen, stable} incorporate text control information into the process of image generation using a language model. In order to reduce the computational load of diffusion models, Stable Diffusion \cite{stable} utilizes VAE to encode images into a latent space for denoising tasks. Our dataset generation pipeline is built upon Stable Diffusion, allowing us to obtain high-quality, diverse images from textual inputs.

\noindent\textbf{Synthetic Segmentation Dataset.}  Prior works have been employed to use synthetic images in constructing training sets to enhance performance in fields such as classification \cite{class2023diff, bansal2023leaving, he2022synthetic, sariyildiz2023fake, trabucco2023effective}, object detection \cite{wu2022synthetic, zhao2023generative, zhao2023flowtext, ni2022imaginarynet, ge2022dall, ge2022neural}, and segmentation \cite{li2023guiding, ma2023diffusionseg, DiffuMask, DATASETDM, datediff, freemask}. The construction of segmentation datasets requires pixel-level annotation of images, posing a significant challenge for dataset generation tasks. Some researches \cite{DATASETDM, freemask} incorporate a portion of real labeled datasets in the construction process to achieve precise segmentation annotations. However, for open-world segmentation, obtaining a sufficiently diverse set of labeled real data is often challenging.
Other works \cite{DATASETDM, DiffuMask} leverage the characteristics of text-to-image generation models to explore the correlation between target categories and images during the image generation process to construct masks. Although freed from dependence on real data, it remains challenging to generate precise masks, especially for complex images. Building upon prior works, our method utilizes Large Language Models (LLM) to generate high-quality text, thereby obtaining images with richer details. We integrate existing open-vocabulary detection models \cite{liu2023grounding} and interactive segmentation models \cite{kirillov2023segany} to acquire high-quality pixel-level annotations.





~\\
~\\

\section{Method}

\begin{figure*}[ht]
    \centering
    \includegraphics[width=0.85 \linewidth]{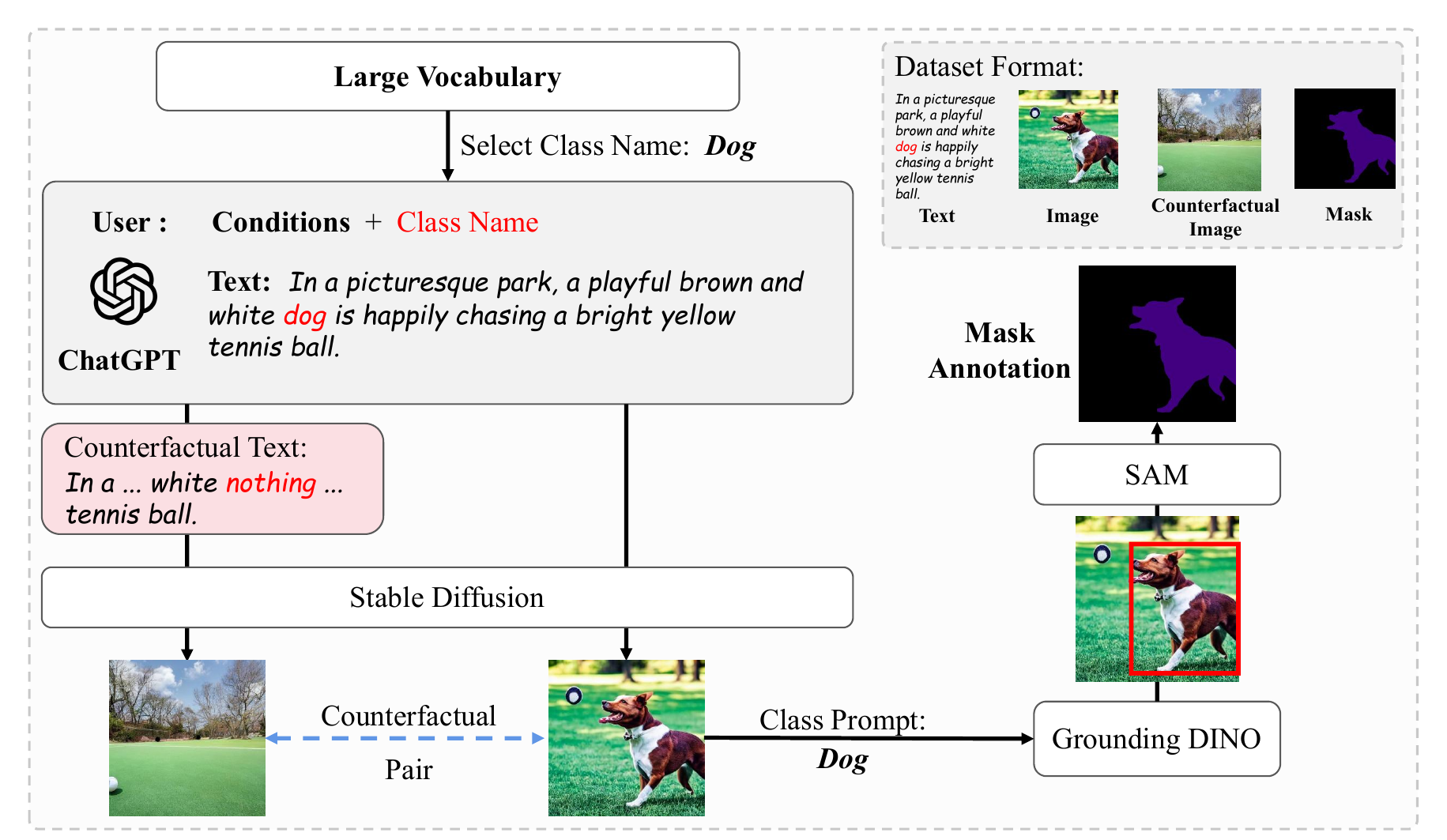}
    \vspace{-1em}
    \caption{{The overall Counterfactual Diffusion-based Generation framework of MagicSeg: creating a large-scale open-world segmentation dataset with pixel-level annotations and counterfactual images for a wide range of categories.}}
    \label{fig:framework_data}
\end{figure*}

\begin{figure*}[ht]
    \centering
    \includegraphics[width=0.85 \linewidth]{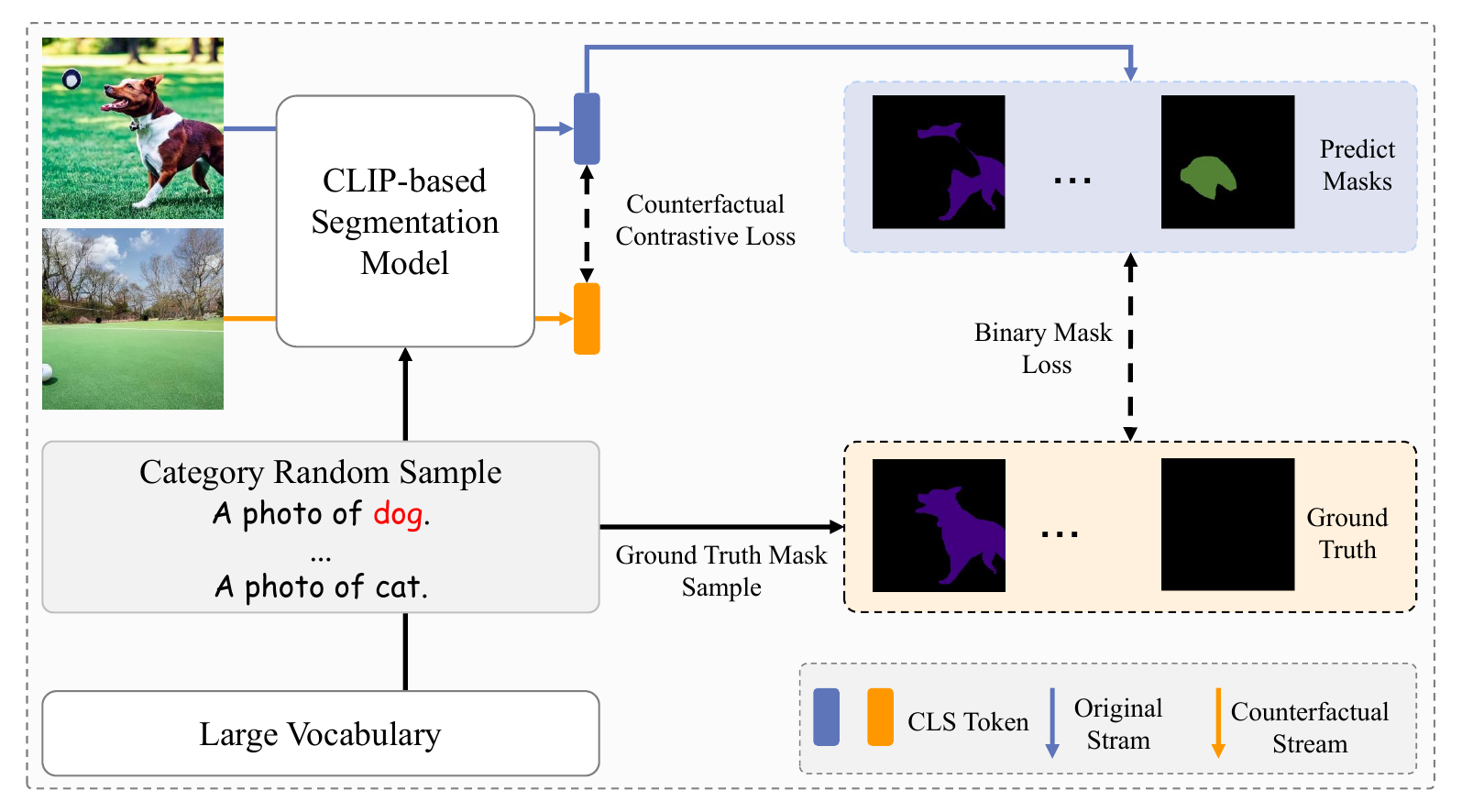}
    \vspace{-1em}
    \caption{{The overall open-world segmentation model training framework of MagicSeg: through the category random sampling strategy and counterfactual contrastive training, we apply the constructed dataset to CLIP-based segmentation model, enhancing its capability in open-world segmentation.}}
    \label{fig:framework_train}
\end{figure*}
The overview of MagicSeg can be seen in Figure \ref{fig:framework_data} and Figure \ref{fig:framework_train}. MagicSeg  consists of two phases. The first stage is Counterfactual Diffusion-based Generation (Section \ref{dataset}), focusing on the process of constructing the segmentation dataset. The second stage is open-world segmentation model training (Section \ref{train}), which explores how to utilize the constructed dataset for training in open-world segmentation.

\subsection{Counterfactual Diffusion-based Generation}
\label{dataset}
As shown in Figure \ref{fig:framework_data}, our dataset construction pipeline is primarily delineated by three sequential stages: text prompt creation, counterfactual image pair generation, and pixel-level mask labeling. To ensure a diverse range of categories within the dataset, we establish a vocabulary $V$ encompassing a multitude of categories. The size of the vocabulary is $N$. Recognizing the criticality of high-quality descriptions for generating images that closely emulate real-world scenarios with intricate details, we exercise caution in managing the number $n$ of categories in one text. Accordingly, we selectively extract one to two class names $C = \{c_1, ..., c_n\}\subset V$ at a time from the vocabulary for text prompt creation. To guide the large language model (\emph{i.e.} ChatGPT) in generating corresponding text $t$, we invoke a set of conditions, such as "The scenarios and details for these examples are as diverse as possible". The details of the text generation conditions are available in  Figure \ref{fig:prompt_app}.

\begin{figure}
    \centering
    \vspace{-1.5em}
    \includegraphics[width=0.8 \linewidth]{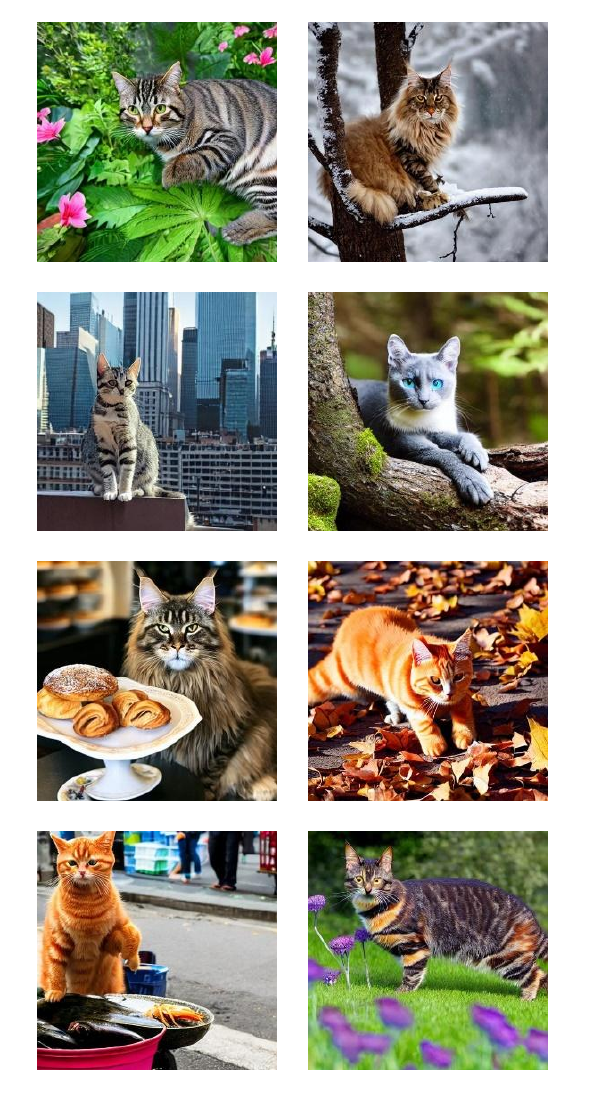}
    \vspace{-1em}
    \caption{Visualization of the diversity of MagicSeg's dataset with class “cat”. It can be seen that MagicSeg can generate diverse images for each category.}
    \vspace{-1em}
    \label{fig:diversity}
\end{figure}

\begin{figure}[htbp]
    \centering
    \includegraphics[width=1.0 \linewidth]{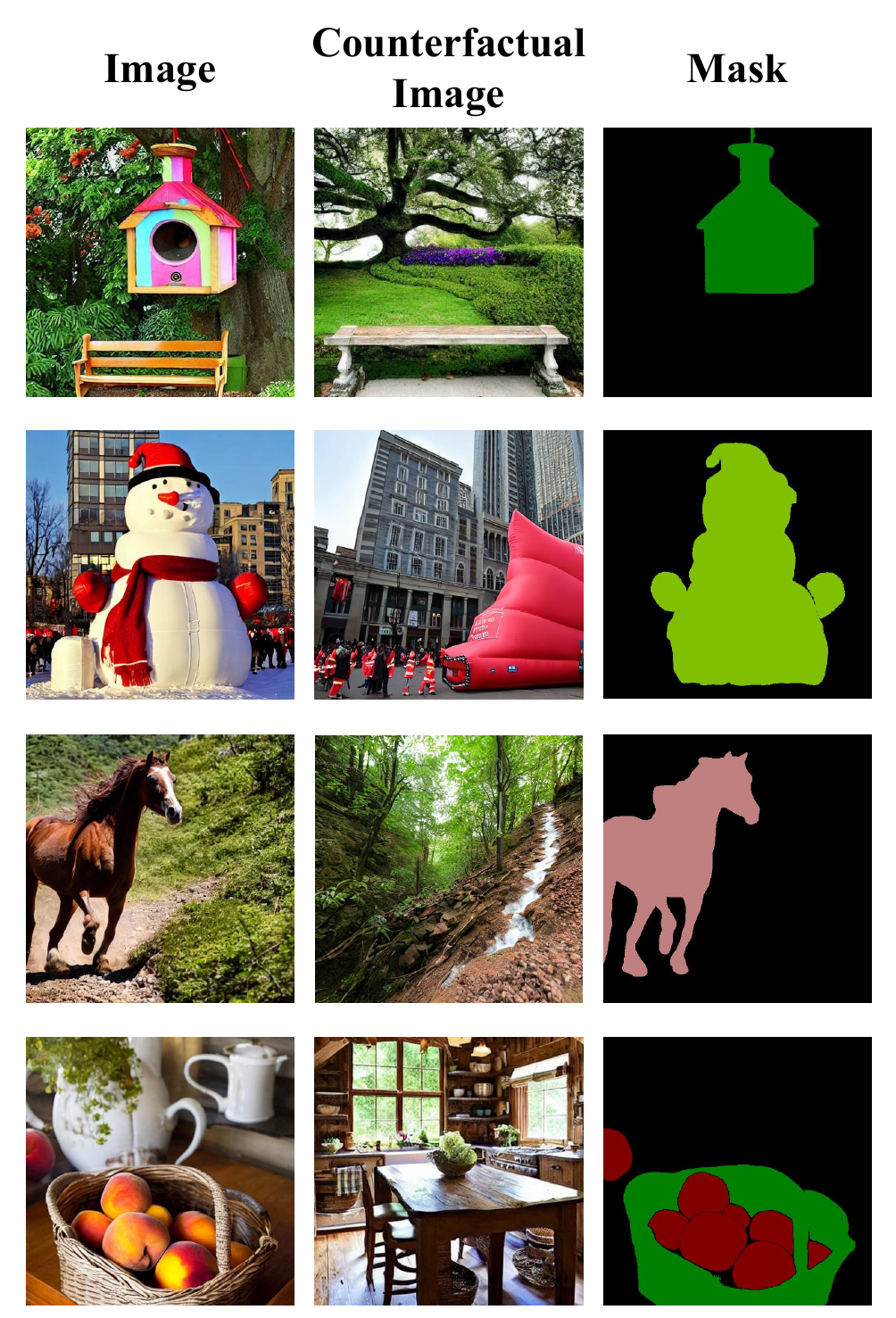}
    \vspace{-1em}
    \caption{Visualization of counterfactual image pairs with mask annotations.}
    \vspace{-1em}
    \label{fig:counterfactual}
\end{figure}

After text prompt creation, we employ Stable Diffusion \cite{stable} to generate images $x$. Because of the high-quality text, we can get diverse images for each class,  which can be seen in Figure \ref{fig:diversity}. Simultaneously, to enhance the object localization capability of the open-world model and overcome noise in the generated image masks, we replace class names in the text $t$ with "nothing" to create counterfactual text $t_{co}$. This substitution creates images $x_{co}$ resembling the originals but lacking the corresponding categories, just as shown in Figure \ref{fig:counterfactual}, forming counterfactual image pairs $(x, x_{co})$. These pairs are then utilised for counterfactual contrastive training in the training phase. For the generated original images $x$, we employ Grounding DINO \cite{liu2023grounding} to identify objects in the images based on the class names $C$ used during generation,  which can exclude images where the target object was not generated at the same time. Subsequently, we use the obtained bounding boxes as prompts for SAM \cite{kirillov2023segany}, an interactive segmentation model to extract masks, thereby achieving pixel-level annotation $Y = \{y_{c_1}, ..., y_{c_n}\} \in \mathbb{R} ^ {n \times H \times W}$ of the images.

With the dataset construction pipeline, we can obtain the synthetic dataset which contains texts, images, masks, and counterfactual images for open-world segmentation.

\subsection{Open-world Segmentation Model Training}
\label{train}
\subsubsection{Category Random Sampling Strategy}

As shown in Figure \ref{fig:framework_train}, our training pipeline utilizes a segmentation model $z$ based on CLIP (\emph{i.e} ZegCLIP \cite{zhou2023zegclip}) to train our segmentation dataset, predicting masks by leveraging the correlation between image features from the image encoder and the text embeddings. However, applying a dataset with a large number of categories to open-world segmentation poses a challenge. Each image $x \in \mathbb{R} ^{D \times H \times W}$, where $D$ denotes the channels, and $H$, $W$ denote the height and width of the image, has a limited number $n$ of categories with masks, most categories in the vocabulary do not appear in the image. During training phase, if we predict masks for all the categories in vocabulary for one image, formulated as
\begin{equation}
    \hat{Y} = z(x, V) \in \mathbb{R}^{N \times H \times W},
\end{equation}
the learning efficiency of the categories which exist in the image will be reduced because of a large number of non-existent categories $(n \ll N)$. Meanwhile, this way could result in a significant computational burden and introduce more noise to affect the performance.

To address this problem, we propose the category random sampling strategy for training. In the training process, for each image $x$, aside from its known categories $C$, we randomly select a certain number of categories from the vocabulary to create a subset $C_{sub} = \{c_1, c_2, ..., c_m\}$ for mask prediction, which can be represented as:
\begin{equation}
    \hat{Y}_{sub} = z(x, C_{sub}) \in \mathbb{R}^{m \times H \times W}.
\end{equation}

It's worth noting that for each image in a batch, we construct a different subset of categories. 
This approach enables the model to predict and learn from random categories during each training iteration, rather than fixed categories, reducing computational load and preventing the model from overfitting to the classes of the constructed dataset.

For the predicted masks $\hat{Y}_{sub} = \{\hat{y}_{c_1}, ..., \hat{y}_{c_m}\}$, considering that, despite having a sufficiently diverse set of categories in the dataset, it is still finite. In order to mitigate the model's suppression of categories it has not learned, following ZegCLIP, we do not use softmax for inter-class suppression. Instead, we utilized sigmoid to independently compute the loss function for masks of different categories. Specifically, we employed the Binary Cross-Entropy (BCE) loss in the type of focal loss and Dice loss to calculate the loss function for predicted masks $\hat{Y}_{sub}$ and ground truth:

\begin{equation}
\begin{split}
        \mathcal{L}_{focal}\!=\!  - \frac{1}{mHW}\!\sum_{i=1}^m\!\sum_{k=1}^{HW}\!\{(1\! -\! \hat{y}_{(c_i, k)})^{\alpha} \! ~~\! y_{(c_i, k)}log(\hat{y}_{(c_i, k)}) \\
    +\ \   \hat{y}_{(c_i, k)}^{\alpha}~~ \!\! (1 \!- \!y_{(c_i, k)})log(1 - \hat{y}_{(c_i, k)})\},
\end{split}
\end{equation}

\begin{equation}
    \mathcal{L}_{Dice} = - \frac{1}{m}\sum_{i=1}^m(1 - \frac{2\Sigma_{k=1}^{HW}\hat{y}_{(c_i, k)}y_{(c_i, k)}}{\Sigma_{k=1}^{HW}\hat{y}_{(c_i, k)}^2 + \Sigma_{k=1}^{HW}y_{(c_i, k)}^2} ),
\end{equation}
where $\alpha$ is the hyper-parameter for the balance of easy and hard samples, $y_{(c_i, k)}$ means ground truth for $c_i$ at the pixel $k$, $\hat{y}_{(c_i, k)}$ means predict mask for $c_i$ at the pixel $k$.
\subsubsection{Counterfactual Contrastive Loss}

As mentioned before, we want to overcome the noise in the pseudo masks that will affect the model's segmentation performance. At the same time, we aim to enable the model to learn class-agnostic object recognition and scene recognition capabilities. To achieve this, MagicSeg uses images that are similar to the original scene but lack the target category as counterfactual images for self-supervised training. By extracting class token $p_{cls}, p_{co}$  from the image encoder of $z$ for both the original and counterfactual images, we calculate cosine similarity for contrastive training:
\begin{equation}
    \mathcal{L}_{cos} = max(0, \frac{p_{cls} \cdot p_{co}}{\left\lVert p_{cls} \right\rVert \left\lVert p_{co} \right\rVert}).
\end{equation}

\subsubsection{Overall Loss Function}
Finally, the total loss of MagicSeg training pipeline can be represented as:
\begin{equation}
    \mathcal{L} = w_1 \mathcal{L}_{focal} + w_2  \mathcal{L}_{Dice} + w_3 \mathcal{L}_{cos},
\end{equation}
where the $w_1, w_2, w_3$ are the weights of each loss function.
~\\
~\\

\section{Experiment}
\subsection{Implementation Details}
\noindent\textbf{Architecture.}
In the counterfactual diffusion-based generation pipeline, {ChatGPT (version GPT-3.5) is employed for text generation, while Stable Diffusion version 1.5 is used for image generation. For pixel-level labeling, we utilized Grounding DINO \cite{liu2023grounding} as the open-vocabulary detector to detect the target object, followed by segmentation using SAM \cite{kirillov2023segany} based on the bounding boxes. During the training phase, MagicSeg utilizes ZegCLIP \cite{zhou2023zegclip}, a segmentation model based on the pre-trained CLIP ViT-B/16} \cite{CLIP} to train our dataset.

\noindent\textbf{Dataset.} For the training dataset, we initially create a vocabulary containing class names from LVIS \cite{gupta2019lvis} and PASCAL VOC \cite{everingham2010pascal}, resulting in a total of 1205 categories. Using this constructed vocabulary, we built a training set of 380k data pairs consisting of text, images, counterfactual images, and pixel-level annotations for open-world segmentation training.
As for the test set, we utilized PASCAL VOC \cite{everingham2010pascal}, PASCAL Context \cite{mottaghi2014role}, COCO \cite{lin2014microsoft}, and LVIS \cite{gupta2019lvis} as our validation sets. These four datasets have 20, 80, 59, and 1203 categories, respectively. And the validation sets include approximately 1.5k, 5k, 5k, and 5k images. Each of these datasets also includes a background class. {To further verify the effectiveness on unseen classes, we conducted evaluation on the OpenImages-v7 dataset  for zero-shot point prediction, which contains approximately 5,000 classes in about 36k images.}

\noindent\textbf{Evaluation Metrics.} For semantic segmentation, we apply mIoU as the evaluation metric to assess the performance of our segmentation models. 
It measure the overlap between predicted and ground truth masks across all categories. 
For background class, we used a threshold of 0.95 to classify it.
For zero-shot point prediction, as the number of categories and the amount of data increase, the computational cost of calculating the mIoU for each full image becomes daunting, following \cite{gem}, 
we calculate the mIoU on sampled points as p-mIoU, the effectiveness of p-mIoU is verified in \cite{benenson2022colouring}. The p-mIoU is highly correlated with mIoU, and through p-mIoU, the segmentation capability of the model can be fully reflected.

\begin{figure*}[!ht]
    \centering
    \includegraphics[width=1.0 \linewidth]{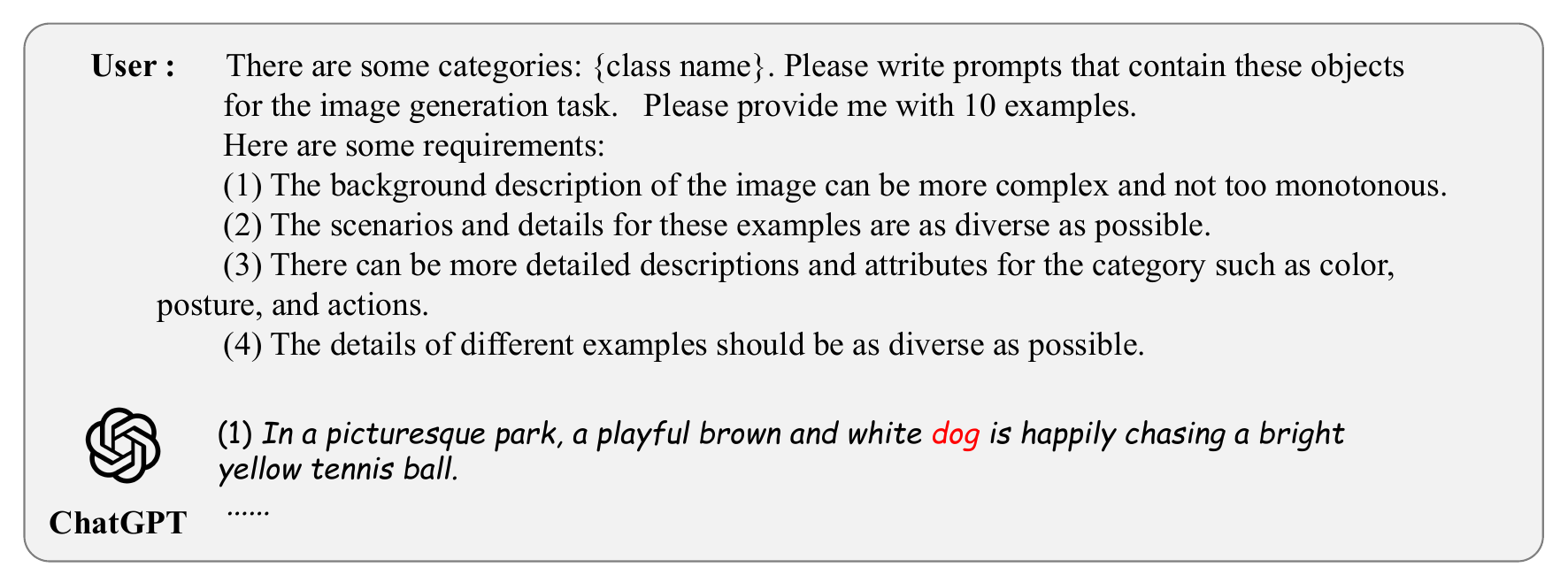}
    \vspace{-2em}
    \caption{{Text prompt generation conditions for ChatGPT. We  design a series of conditions to control the diversity of generated text.}}
    \label{fig:prompt_app}
    \vspace{-1em}
\end{figure*}

\noindent\textbf{Text Generation Conditions}
Our conditions for ChatGPT to generate text prompts can be seen in Figure \ref{fig:prompt_app}. We initially select a certain number of class names to provide to ChatGPT and generate prompts based on specific requirements. 
These requirements are primarily considered from the following perspectives: 
\begin{itemize}
    \item To ensure that the generated images do not let the generated objects dominate the majority of the image, causing the background of the generated image to be overly monotonous, we require the generated text to provide sufficient descriptions of the background while not affecting the main status of the target category. 
    \item In order for the same category to appear in different environments and to ensure that the generated content has sufficient diversity, we require the scenes generated by ChatGPT to be as diverse as possible and have a certain level of detail.
    \item In order to diversify the attributes of the target category, thereby enabling the model to learn segmentation with generalization and reduce the occurrence of overly monotonous generated categories.
\end{itemize}
\noindent\textbf{Training Details.} We train our model on 16 Tesla V100s with a batch size of 64 and the image resolution is $512 \times 512$. During training, the size $m$ of $C_{sub}$ for each image to predict masks is set to 100. The optimizer is set to Adam with a learning rate of 0.0002, and the pre-trained weights of the image encoder and text encoder of CLIP \cite{CLIP} are frozen. Follow \cite{zhou2023zegclip}, the weight for the focal loss $w_1$ is set to 100, while the weights for dice loss and counterfactual loss $w_2$ and $w_3$ are both set to 1.
\begin{table}[ht]
    \caption{Performance comparison with baseline Grounding DINO \cite{liu2023grounding} + SAM \cite{kirillov2023segany} (Grounded SAM) on PASCAL VOC and LVIS.}
    \centering
    \begin{tabular}{c|c|c}
    \hline
    \multirow{2}{*}{Method}& \multicolumn{2}{c}{Val (mIoU(\%))}  \\\cline{2-3}
          & PASCAL VOC & LVIS \\\hline
    Grounding DINO + SAM & 61.8 & 0.5 \\
    MagicSeg & 62.9 & 4.9 \\\hline
    \end{tabular}

    \label{tab:baseline}
\end{table}
\subsection{Comparison}
\noindent\textbf{Baseline.} Since we get our annotations from Grounded SAM (Grounding DINO \cite{liu2023grounding} + SAM \cite{kirillov2023segany}), we compared our performance with Grounded SAM (Grounding DINO \cite{liu2023grounding} + SAM \cite{kirillov2023segany}) as baseline, as shown in the Table \ref{tab:baseline}. It can be observed that on Pascal VOC, our performance is superior to Grounded SAM (62.9\% \emph{vs.} 61.8\%). On the LVIS dataset, as the number of categories increases and the situation where the categories contained in the image are unknown, the performance of Grounded SAM significantly declines, while we still demonstrate superior performance compared to the baseline (4.9\% \emph{vs.} 0.5\%).

\noindent\textbf{Open-World Segmentation.} We do a comprehensive evaluation of MagicSeg against open-world semantic segmentation methods. As shown in Table \ref{tab:open-world}, we summarize the comparison results on PASCAL VOC, PASCAL Context, and COCO.

Compared to the open-world semantic segmentation methods without mask supervision, the results show the superiority of our approach, confirming its effectiveness. To demonstrate the ability of our method in open-world segmentation, we first train our model with MagicSeg-20k-20c, a dataset containing only 20 classes from PASCAL VOC. Despite the limited number of categories, our model exhibits strong segmentation capabilities on PASCAL Context and COCO, closely matching the performance of some previous open-world segmentation models like \cite{Mixreorg, viewco} (21.3\% \emph{vs.} 23.9\%, 20.5\% \emph{vs.} 23.5\%). {The full dataset, MagicSeg-380k-1205c, demonstrates robust performance, outperforming previous open-world segmentation models on PASCAL VOC (62.9\% \emph{vs.} 59.1\%) and exhibiting strong performance on COCO (40.2\% \emph{vs.} 32.0\%).} Despite PASCAL Context containing a significant number of stuff classes, while our method primarily focuses on thing classes, our model is still comparable to the previous models (26.7\% \emph{vs.} 34.5\%). 

Compared to the method using real data with mask annotations, there is still a certain performance gap between MagicSeg and them. Nonetheless, MagicSeg notably reduces the disparity between open-world segmentation methods and those reliant on annotated data (62.9 \% \emph{vs.} 82.7 \%,  52.4\% \emph{vs.} 82.7\%). In some cases, MagicSeg even surpasses the previous method (40.2 \% \emph{vs.} 36.1\%).  
Our dataset is limited to a maximum of two categories per image. However, this constraint does not compromise the model's segment capability, as the model learns fine-grained alignment—specifically, aligning local image regions with one of the 100 target categories.
\begin{table*}[!t]
    \caption{Performance comparison on PASCAL VOC, PASCAL Context, and COCO with open-world semantic segmentation methods."20k" and "380k" indicate the size of the dataset. "20c" represents that the dataset is composed of only 20 categories of VOC, while "1205c" means we use the full vocabulary $V$ to design the dataset. The superscript $^*$ denotes the results are from \cite{dong2022maskclip}. Different from MagicSeg, ODISE \cite{xu2023open} and OpenSeg \cite{ghiasi2022scaling} use real data with manual annotations. \sethlcolor\hl{PT stands for pre-training, while FT stands for fine-tuning.}
    }
    \centering
    \begin{tabular}{lcc|c|c|c} 
    \toprule
     \multicolumn{3}{c|}{ Pre-training } & \multicolumn{3}{c}{Validation (mIoU (\%))} \\\hline
     Method & Dataset & Supervision  & \makecell{PASCAL VOC} & \makecell{PASCAL Context} & COCO  \\\hline 
    \color{gray}ODISE\cite{xu2023open} & \color{gray}COCO & \color{gray}mask & \color{gray}82.7 & \color{gray}55.3 & \color{gray}52.4 \\
    \color{gray}OpenSeg\cite{ghiasi2022scaling} & \color{gray}COCO & \color{gray}mask & \color{gray}- & \color{gray}42.1 & \color{gray}36.1 \\
    
    \cline{1-6}
     CLIP-MAE$^*$ \cite{dong2022maskclip}   & LAION-20M     & text \& self    & -     & $16.8$ & 11.8 \\
     MaskCLIP \cite{dong2022maskclip}       & LAION-20M     & text \& self      & -     & $17.7$ & 11.8 \\
     ViewCo \cite{viewco}                   & CC12M+YFCC        & text \& self      & 52.4  & 23.0   & 23.5
    \\\cline{1-6}
    MaskCLIP \cite{zhou2021denseclip}       & CLIP-400M     & text              & -     & $21.7$ & -\\
    CLIP$^*$ \cite{CLIP}                    & LAION-20M     & text              & -     & $13.5$ &  8.2 \\
    GroupViT\cite{xu2022groupvit}           & CC12M+YFCC    & text              & 51.2 & $22.3$ &20.9 \\
    MixReorg\cite{Mixreorg}           & CC12M    & text              & 47.9 & 23.9 &21.3 \\
    TCL\cite{cha2023learning} & CC12M & text & 55.0 & 30.4 & 31.6 \\
    SegCLIP \cite{luo2023segclip} & CC3M+COCO& text & 52.6 & 24.7 & 26.5 \\
    CLIPpy \cite{ranasinghe2023perceptual} &HQITP-134M & text &52.2  &-     & \underline{32.0}  \\
    OVS \cite{xu2023learning} & CC4M & text & 53.8 & 20.4 & 25.1 \\
    {CLIP Surgery} \cite{li2023clipsur} & CLIP-400M & text & 41.2 & 30.5 & - \\
    {SCLIP}\cite{wang2024sclip} &CLIP-400M& text& \underline{59.1} & 30.4 & 30.5 \\
    {GEM}\cite{gem} & MetaCLIP-400M & text & 46.8 & \underline{34.5} & -\\
    \cline{1-6}
    MagicSeg (Ours)           & {CLIP-400M(PT)} + MagicSeg-20k-20c(FT)    &  synthetic             & \textbf{64.3} & 21.3 & 20.5\\
    MagicSeg (Ours)          & {CLIP-400M(PT)} + MagicSeg-380k-1205c(FT)    &  synthetic            & \textbf{62.9} & 26.7 & \textbf{40.2}\\
    
    \bottomrule
    
    \end{tabular}
    \label{tab:open-world}
\end{table*}

\begin{table}[ht]
    \centering
    \caption{{Comparison on zero-shot point prediction on OpenImages-V7.}}
    \begin{tabular}{c|c}
    \hline
       \textbf{Method} & \textbf{p-mIoU (\%)} \\\hline
       OVSeg\cite{liang2023open} & 22.5 \\
       SegCLIP\cite{luo2023segclip} & 32.1 \\
        GroupViT\cite{xu2022groupvit} & 36.5 \\
        CLIP\cite{CLIP} & 27.6 \\
        MaskCLIP\cite{zhou2021denseclip} & 42.0  \\\hline
        Groundingdino + SAM\cite{liu2023grounding, kirillov2023segany} & 53.3 \\
        GEM-SAM-MetaCLIP\cite{gem} & 55.2 \\ 
        \hline\hline
        MagicSeg(Ours) & 45.6 \\\hline
    \end{tabular}
    
    \label{tab:open}
\end{table}

\begin{table*}[!t]
    \caption{Performance comparison on PASCAL VOC and COCO with works for synthetic dataset. "R50" and "R101" mean resnet50 and restnet101 respectively. Our dataset generation pipeline is primarily designed for open-world segmentation without specific settings for different datasets.
    }
    \centering
    \begin{tabular}{c|ccc|ccc} 
    \toprule
     \multirow{2}{*}{Segmenter} & \multicolumn{3}{c}{PASCAL VOC} & \multicolumn{3}{c}{COCO} \\
    \cline{2-5} \cline{6-7}
          & Dataset & Class Number & Val (mIoU(\%))    & Dataset & Class Number & Val (mIoU (\%))\\\hline
        \color{gray}\makecell{DeepLabV3  (R50)} & \color{gray}\multirow{3}{*}{\makecell{VOC's training \\(real: 11.5k)}} &  \color{gray}20  & \color{gray}77.4  & \color{gray}\multirow{3}{*}{\makecell{COCO's training \\(real: 118k)}} &  \color{gray}80  & \color{gray}48.9  \\
        \color{gray}\makecell{DeepLabV3  (R101)} &  &  \color{gray}20  & \color{gray}79.9  &  &  \color{gray}80  & \color{gray}54.9  \\
        \color{gray}\makecell{Mask2Former  (R50)} &  &  \color{gray}20  & \color{gray}77.3  &  &  \color{gray}80  & \color{gray}57.8  \\\hline
       Mask2Former  (R50) &  \makecell{DiffuMask \cite{DiffuMask}\\(synthetic: 60k)} & 20 & 57.4 & - & - & - \\\hline
        DeepLabV3  (R50) &  \multirow{3}{*}{\makecell{Dataset Diffusion \cite{datediff} \\(synthetic: 40k)}} & 20 & 61.6 & \multirow{3}{*}{\makecell{Dataset Diffusion \cite{datediff} \\(synthetic: 80k)}} & 80 & 32.4 \\
         DeepLabV3  (R101) &   & 20 & 64.8 &  & 80 & 34.2 \\
        Mask2Former  (R50) &   & 20 & 60.2 &  & 80 & 31.0 \\\hline\hline
        
        DeepLabV3  (R50) &  \multirow{3}{*}{\makecell{MagicSeg (Ours)\\(synthetic: 20k)}} & 20 & 62.9 & \multirow{3}{*}{\makecell{MagicSeg (Ours)\\(synthetic: 40k)}} & 80 & 32.6 \\
         DeepLabV3  (R101) &   & 20 & \textbf{65.3} &  & 80 & 34.0 \\
        Mask2Former  (R50) &   & 20 & 59.3 &  & 80 & 30.5 \\
        
        
    ZegCLIP &  \makecell{MagicSeg (Ours)\\(synthetic: 380k)} & \textbf{1205} & 62.9 & {\makecell{MagicSeg (Ours)\\(synthetic: 380k)}} & \textbf{1205} & \textbf{40.2} \\    
    
    \bottomrule
    
    \end{tabular}
    \label{tab:gen}
\end{table*}

\noindent{\noindent\textbf{Zero-Shot Point Prediction.} 
As shown in Table \ref{tab:open}, we conducted a comprehensive comparison of MagicSeg with previous open-world segmentation methods on the zero-shot point prediction task. The OpenImages-v7 dataset, which contains a significantly larger number of classes than our training set, fully demonstrates the model's open-world segmentation capabilities.
It can be seen that, although our method has a certain gap compared to the method based on SAM, MagicSeg still demonstrates competitive zero-shot capability among a range of open-world segmentation methods. We attribute this gap to the fact that these baselines benefit from extensive pre-training on massive datasets with significantly broader category coverage compared to our synthetic training set.
}

\noindent\textbf{Synthetic Dataset.} In Table \ref{tab:gen}, MagicSeg is  compared with several extant methods \cite{DiffuMask, datediff} dedicated to the generation of semantic segmentation datasets.  

For the consistency of the experimental setting and to validate the effectiveness of MagicSeg’s dataset, we first compare our dataset under the same backbones and with the same number of categories. It can be seen that under the same conditions, our dataset achieves performance that equals or surpasses that of previous methods (65.3\% \emph{vs.} 64.8\%, 34.0\% \emph{vs.} 34.2\%). 

It is worth noting that, in contrast to other methods, MagicSeg's strength lies in not requiring the construction of datasets tailored to different downstream tasks. During the evaluation phase on the PASCAL VOC, our full synthetic dataset, constructed with a significantly larger number of categories than the target dataset (1205 \emph{vs.} 20), manifests a performance that is remarkably commendable (62.9\%). MagicSeg performs competitively with existing methods. Additionally, the versatility of our segmentation model is demonstrated through its applicability to different datasets without necessitating new dataset construction. Notably, on COCO, MagicSeg eclipses methods expressly engineered for dataset construction based on COCO categories (40.2\% \emph{vs.} 34.2\%). The experimental findings robustly attest to the efficacy of the dataset we have constructed and underscore the universal applicability of MagicSeg across a series of downstream segmentation tasks.

\subsection{Ablation Study}
\begin{table*}[!ht]
\begin{minipage}[b]{.48\linewidth}
    \caption{Ablation study of text prompt on COCO. Simple Prompt means "a photo of \{\textit{class name}\}". }
    \centering
    \begin{tabular}{c|c|c|c}
    \hline
    Method  & Counterfactual & Dataset Size  & mIoU (\%) \\\hline
    Simple Prompt &  \XSolidBrush & \multirow{4}{*}{40k} & 28.6 \\
    Simple Prompt &  \Checkmark &   & 27.5 \\
    MagicSeg &  \XSolidBrush &   & 36.0 \\
    MagicSeg &  \Checkmark &   & \textbf{37.2} \\    
    \hline
    \end{tabular}
    \label{tab:prompt}
    \end{minipage}
\begin{minipage}[b]{.48\linewidth}
    \caption{Ablation Study of Counterfactual Contrastive Loss and the number $m$ of subset $C_{sub}$ for Category Random Sampling Strategy on COCO.
    }
    \centering
    \begin{tabular}{cc|c}
    \hline
    Counterfactual & $m$ of $C_{sub}$  & mIoU (\%)  \\\hline
    \XSolidBrush & 1 & 7.7 \\
          \XSolidBrush & 1205 & 31.4 \\
          \XSolidBrush & 100 & 38.1 \\
          \Checkmark   &  100 & \textbf{40.2} \\\hline
    \end{tabular}
    \label{tab:ablation}
\end{minipage}
\end{table*}

\noindent\textbf{Simple Text Prompt \emph{vs.} MagicSeg's Text Prompt.} We study the text prompt design method, examples can be seen in Table \ref{tab:text}. Since counterfactual images may have different effects on different types of prompts, we first compare the situation without counterfactual contrastive learning. It can be seen that MagicSeg's text prompt design method outperforms the simple text prompt (36.0 \% \emph{vs.} 28.6 \%), which uses "a photo of " with class names. Due to MagicSeg using ChatGPT to generate texts according to conditions, we can obtain texts with rich description details, thus obtaining diverse images, which is helpful for the model to improve the ability of segmentation. 

\begin{figure*}[!ht]
    \centering
    \includegraphics[width=0.9 \linewidth]{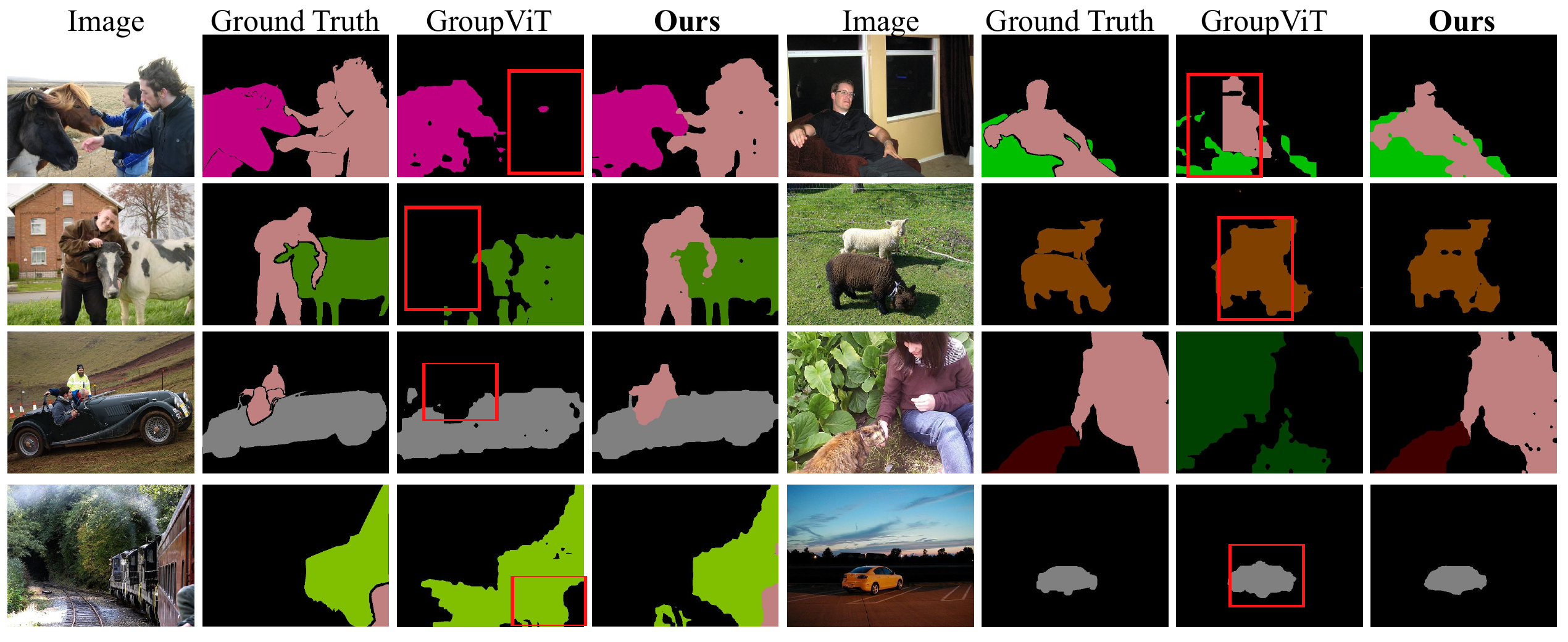}
    \vspace{-1em}
    \caption{Semantic segmentation results on PASCAL VOC. Compared to GroupViT, MagicSeg handles details such as mask edges more effectively. }
    \vspace{-1em}
    \label{fig:vis_voc}
\end{figure*}

\begin{figure*}[!ht]
    \centering
    \includegraphics[width=0.9 \linewidth]{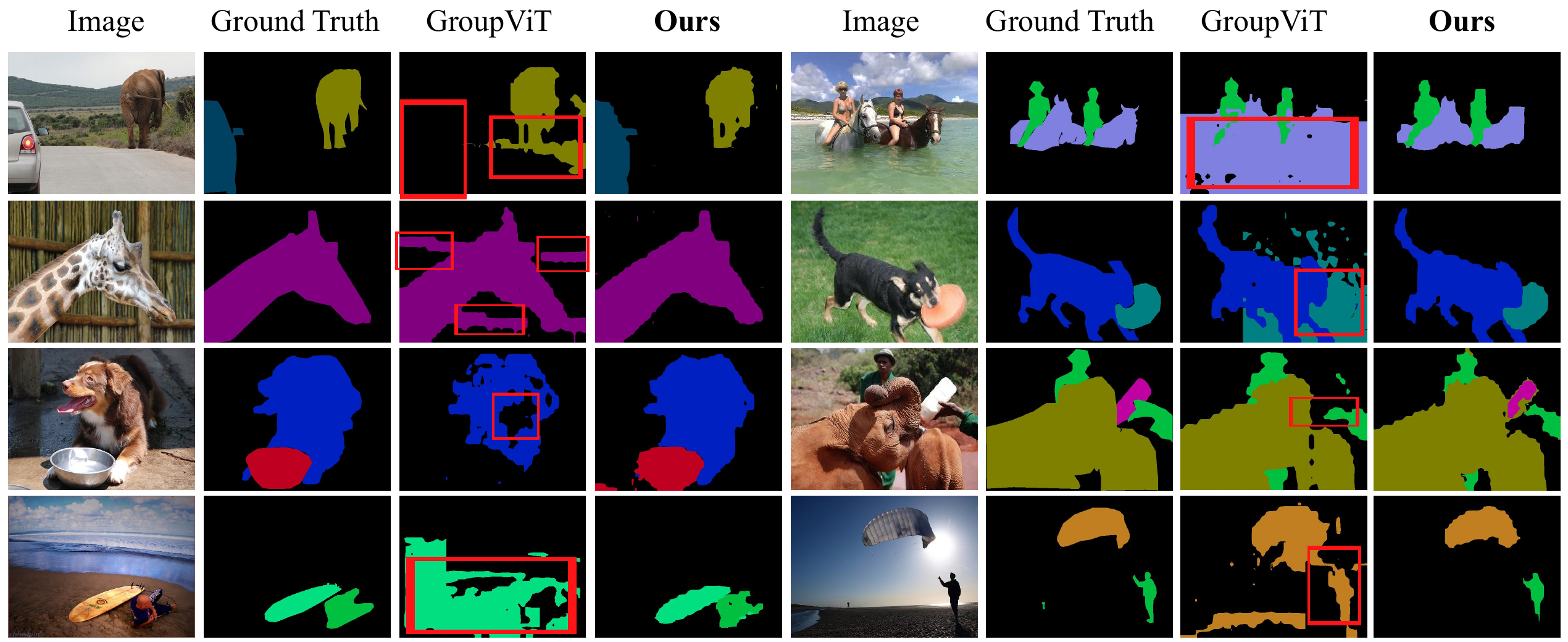}
    \vspace{-1em}
    \caption{Semantic segmentation results on COCO. Compared to GroupViT, MagicSeg shows a clear advantage in fine-grained category recognition and detail handling. The quality of the masks has significantly improved, reducing the gap with supervised training.}
    \label{fig:vis_coco}
\end{figure*}

\begin{table}[ht]
\caption{Examples of simple text prompt and MagicSeg's prompt.}
    \centering
    \begin{tabular}{l|c|c}
    \hline
     & prompt & counterfactual prompt \\\hline
         \textbf{Simple}&  A photo of \underline{\textbf{bus}}. & A photo of \underline{\textbf{nothing}}.\\\hline
         \textbf{ Ours}& \makecell{In a peaceful countryside \\
                                    setting, a yellow school \\ 
                                    \underline{\textbf{bus}} stops by a quaint \\ 
                                    picket fence, .....} 
                                & \makecell{In a peaceful countryside \\
                                    setting, a yellow school \\
                                    \underline{\textbf{nothing}} stops by a quaint \\
                                    picket fence, .....}\\\hline
    \end{tabular}
    \label{tab:text}
\end{table}

Additionally, we study the impact of counterfactual contrastive training on different types of prompts. As shown in Table \ref{tab:prompt}, we use counterfactual images for the datasets obtained from the simple prompt and MagicSeg's prompt for contrastive training. The example of counterfactual prompt can be seen in Table \ref{tab:text}. It can be observed that the counterfactual contrastive training still works effectively for our prompt (37.2\% \emph{vs.} 36.0 \%), but it has a counterproductive effect on the data from the simple prompt (27.5\% \emph{vs.} 28.6).  For the simple prompt, the counterfactual images that come from "A photo of nothing" lose a large amount of information, rendering the counterfactual contrastive training ineffective. In contrast, MagicSeg's prompt still contains a large amount of text describing the background and other information in the image for counterfactual images, thus playing a strong role in contrastive training.

\noindent\textbf{The Number $m$ of the subset $C_{sub}$ for Category Random Sampling Strategy.} In Table \ref{tab:ablation}, we conduct the ablation study on the number $m$ of samples for the category random sampling strategy. It can be observed that the optimal sampling number $m$ is 100 (38.1\% \emph{vs.} 31.4\% \emph{vs.} 7.7\%). It can be seen that when the sampling number is only 1, which means each image is trained on only one known category, it will lead to a lack of knowledge about negative examples that result in negatively impacting the model's learning (7.7\%). Another case is when $m = 1205$ (i.e., training with all categories from the vocabulary), compared to the case with $m = 1$, performance is significantly improved (31.4\% \emph{vs.} 7.7\%). For each image, segmentation training is performed not only on existing categories but also on non-existing categories. However, due to the large number of categories in the vocabulary and the limited number of categories present in each image, predicting segmentation for all categories in the vocabulary would result in using a large number of all-zero masks as ground truth, suppressing the model's learning of known categories in the image.

When $m = 100$, we limit the number of non-existing categories for training to a reasonable level, which is more conducive to the model's learning (38.1\% \emph{vs.} 31.4\%). Additionally, using the category random sampling strategy effectively reduces the computational burden. Through the category random sampling strategy, each training iteration involves different categories, helping mitigate biases toward known categories in the vocabulary $V$.

\noindent\textbf{Ablation Study of Counterfactual Contrastive Loss.} As shown in Table \ref{tab:ablation}, we conduct an ablation study focusing on the counterfactual contrastive loss for the full dataset, revealing its pivotal role in the segmentation model (40.2\% \emph{vs.} 38.1\%). The absence of categories in the counterfactual images for contrastive learning enhances the model's ability to recognize and locate segmented objects. In contrast to meticulously labeled object masks, counterfactual images impart segmentation capabilities to the model without reliance on explicit pixel-level annotations, thereby introducing supplementary information. This augmentation proves instrumental in assisting  the model's segmentation learning and mitigating the influence of noise inherent in pixel-level pseudo masks.
\subsection{Visualization}
\noindent\textbf{Qualitative Results.} As shown in Figure \ref{fig:vis_voc} and Figure \ref{fig:vis_coco}, we illustrate semantic segmentation results predicted by GroupViT \cite{xu2022groupvit} and MagicSeg to demonstrate the effectiveness of our method. MagicSeg demonstrates impressive segmentation across various categories. Compared to GroupViT, MagicSeg exhibits notable segmentation capabilities for images containing multiple categories and effectively captures fine details, even for small objects.

On one hand, MagicSeg demonstrates superior performance when dealing with multiple objects and multiple categories. On the other hand, our
method exhibits better performance in mask quality, reducing issues such as edge aliasing. With to the use of a large amount of synthetic datasets with counterfactual images and masked labels, the trained model can better address above issues, benefiting from more granular information guidance. This helps narrow the gap between open-world semantic segmentation and labeled semantic segmentation.
\begin{figure*}[ht]
    \centering
    \includegraphics[width=1.0 \linewidth]{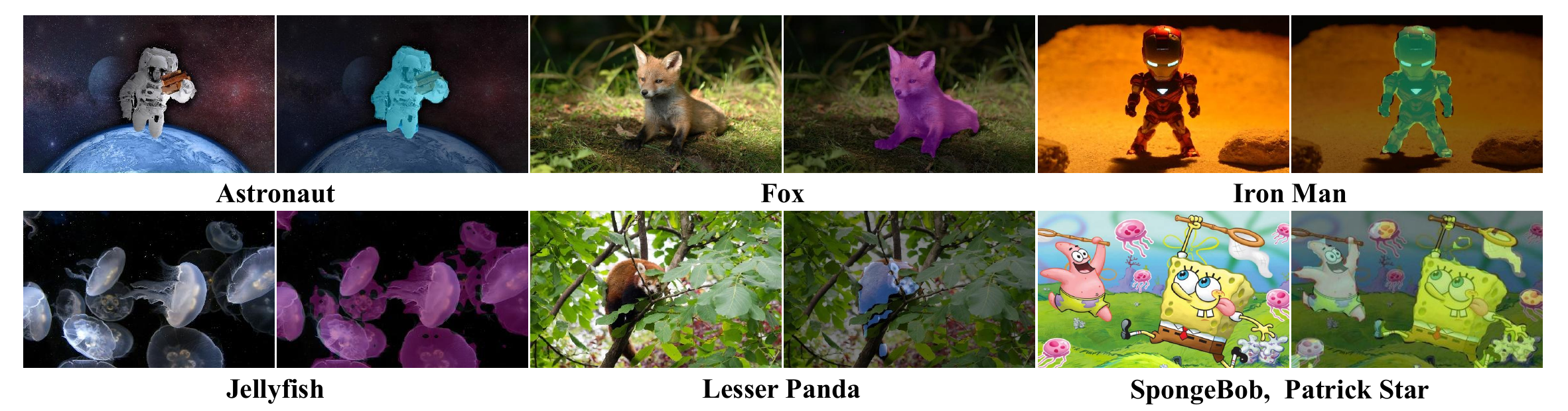}
    \vspace{-2em}
    \caption{Semantic segmentation results on images from network with rare classes.}
    \label{fig:openseg}
\end{figure*}


\noindent\textbf{Open-world Segmentation.} In Figure \ref{fig:openseg}, we visualize the open-world segmentation results predicted by MagicSeg. 
It can be seen that MagicSeg can segment objects for unseen categories, demonstrating the robust segmentation capability of our approach in open-world segmentation.

~\\
~\\
\section{Discussion}
\noindent\textbf{Conclusion. }  We propose  MagicSeg to design a quadruplet dataset for open-world segmentation, including pixel-level annotations and counterfactual images. In addition, we introduce a category random sampling strategy  and counterfactual contrastive training to leverage our dataset in vision-language models, enhancing the model's capability for open-world segmentation.

\noindent\textbf{Limitation. } MagicSeg presents two challenges that require further exploration. Firstly, in each generated image, there is potential to uncover masks for unknown categories in addition to the known ones. Secondly, our approach is constrained by the quality of image generation, which in turn affects the quality of pixel-level labels.

  \section*{Acknowledgments}
This work was supported by the National Key Research and Development Program of China (Grant No.2024YFE0203100), the Scientific Research Innovation Capability Support Project for Young Faculty (No.ZYGXQNJSKYCXNLZCXM-I28), the National Natural Science Foundation of China (NSFC) under Grants No.62476293 and No.62372482, the General Embodied AI Center of Sun Yat-sen University, and in part by the Major Key Project of the PCL (Grant No.PCL2025A17).
~\\
~\\

\ifCLASSOPTIONcaptionsoff
  \newpage
\fi




\bibliographystyle{IEEEtran}
\bibliography{main}

\begin{IEEEbiography}[{\includegraphics[width=1in,height=1.2in,clip,keepaspectratio]{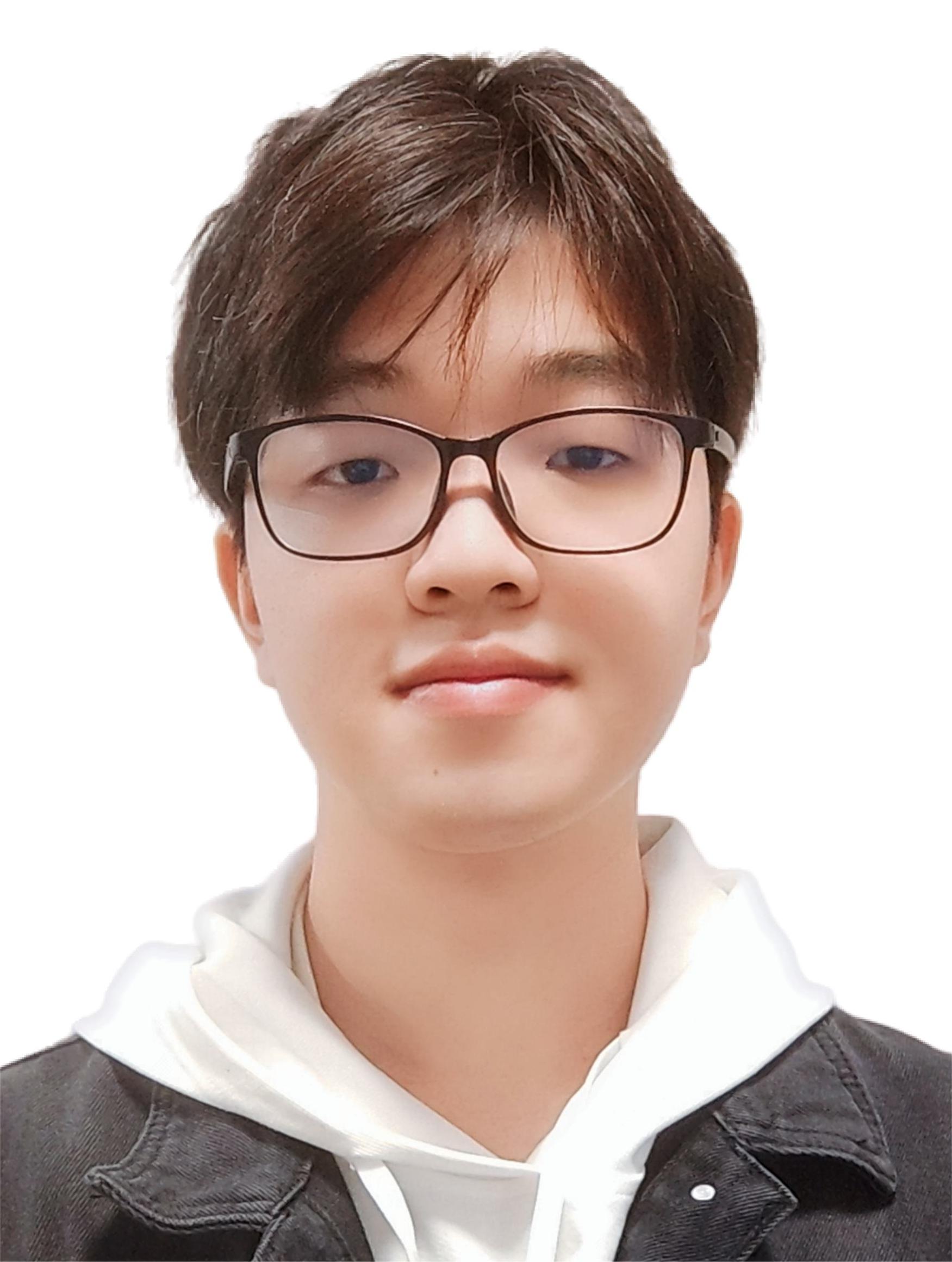}}]{Kaixin Cai}  is currently a postgraduate student at the School of Intelligent Engineering, Sun Yat-Sen University, Shenzhen, China. His
research interests include Segmentation, multimodal learning, and image editing.
\end{IEEEbiography}
\begin{IEEEbiography}[{\includegraphics[width=1in,height=1.2in,clip,keepaspectratio]{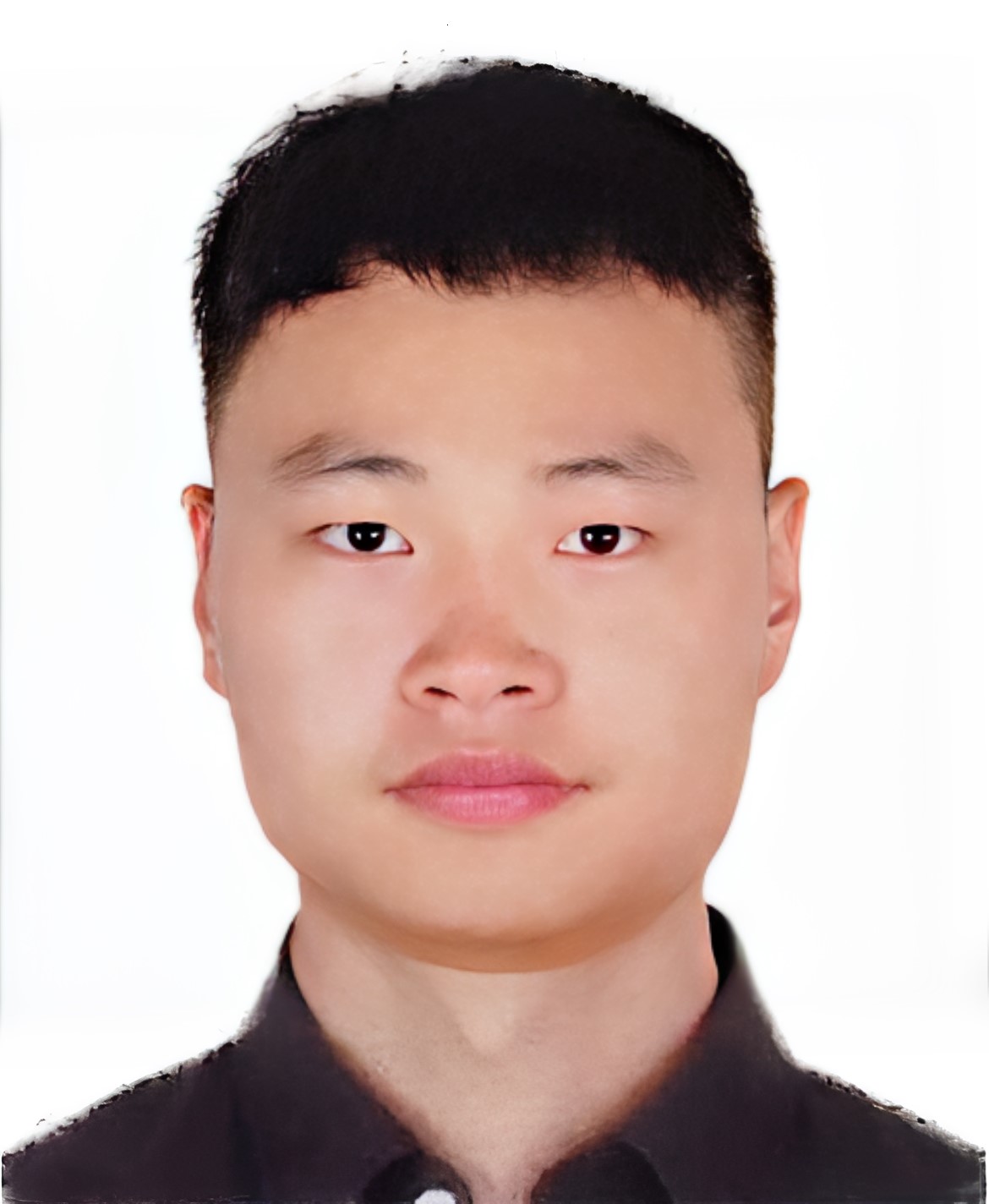}}] { Pengzhen Ren} is currently a Postdoc Research Fellow at the School of Intelligent Engineering, Sun Yat-Sen University. He received his Ph.D. in Software Engineering from Northwestern University in China in 2022. He has published
many papers in top AI conferences and journals such as CVPR, ICLR, AAAI, IJCAI, TPAMI,
and TNNLS. Currently, he mainly focuses on the fields of vision-language pre-training, segmentation, multimodal learning, and visual selfsupervision.     \end{IEEEbiography}

\begin{IEEEbiography}[{\includegraphics[width=1in,height=1.2in,clip,keepaspectratio]{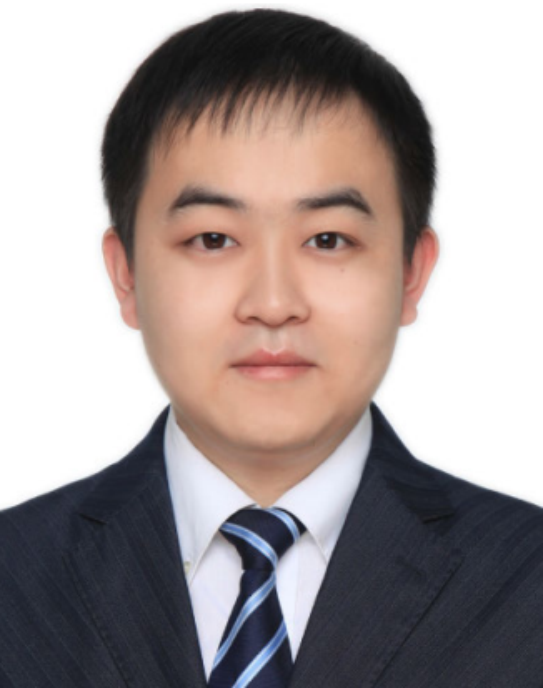}}]{ Jianhua Han}  received bachelor’s and master’s degrees from Shanghai Jiao T ong University, Shanghai, China, in 2016 and 2019, respectively. He is currently a Researcher with the Huawei Noah’s Ark Lab, Huawei T echnologies Company Ltd., Shenzhen, China. His research interests lie primarily in deep learning and computer vision.\vspace{-2em} \end{IEEEbiography}\vspace{-1em}
\begin{IEEEbiography}[{\includegraphics[width=1in,height=1.2in,clip,keepaspectratio]{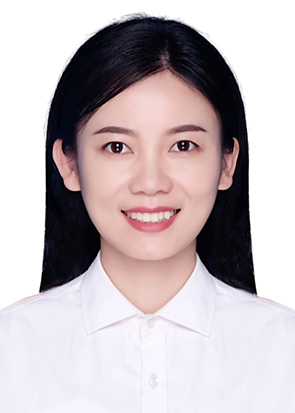}}]{Yi Zhu} is currently a Research Scientist in Huawei Noah’s Ark Laboratory. She received the Ph.D. degree in computer science with the School of Electronic, Electrical, and Communication Engineering, University of Chinese Academy of Sciences,
Beijing, China. Her research interests include scene understanding, weakly supervised learning and vision-language pretraining.
\vspace{-2em}\end{IEEEbiography}          \vspace{-1em}             
\begin{IEEEbiography}[{\includegraphics[width=1in,height=1.2in,clip,keepaspectratio]{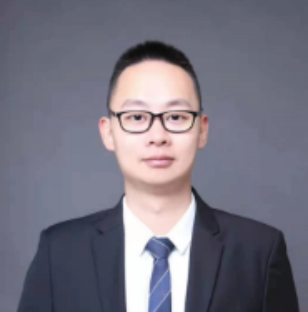}}]{Hang Xu} is currently a senior researcher in Huawei Noah Ark Lab. He received his BSc from Fudan University and Ph.D from Hong Kong University in Statistics. His research interests include multi-modality learning, machine learning, object detection, and AutoML. He has published 70+ papers in T op AI conferences: NeurIPS, CVPR, ICCV , AAAI, and some statistics journals, e.g., CSDA, and Statistical Computing.
\vspace{-2em}\end{IEEEbiography}     \vspace{-1em}                                                                          
\begin{IEEEbiography}[{\includegraphics[width=1in,height=1.2in,clip,keepaspectratio]{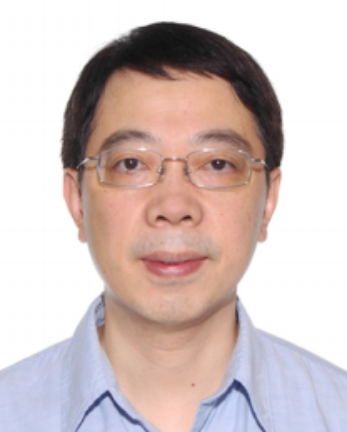}}] {Jianzhuang Liu} (Senior Member, IEEE) received the PhD degree in computer vision from The Chinese University of Hong Kong, in 1997. From 1998 to 2000, he was a research fellow at Nanyang T echnological University, Singapore. From 2000 to 2012, he was a post-doctoral fellow, an assistant professor, and an adjunct associate professor at The Chinese University of Hong Kong, Hong Kong. In 2011, he joined the Shenzhen Institute of Advanced T echnology, University of Chinese Academy of Sciences, Shenzhen, China, as a professor. He was a principal researcher at Huawei Company from 2012 to 2023. He has authored more than 200 papers in the areas of computer vision, image processing, deep learning, and AIGC.
\vspace{-2em}     \end{IEEEbiography}                                                                
\begin{IEEEbiography}[{\includegraphics[width=1in,height=1.2in,clip,keepaspectratio]{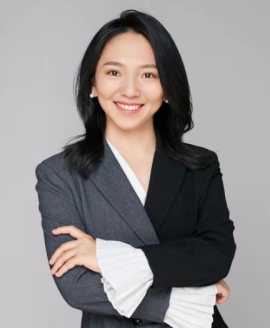}}]{Xiaodan Liang} is currently an Associate Professor at Sun Y at-sen University. She was a postdoc researcher in the machine learning department at Carnegie Mellon University, working with Prof. Eric Xing, from 2016 to 2018. She received her PhD degree from Sun Y at-sen University in 2016, advised by Liang Lin. She has published several cutting-edge projects on human-related analysis, including human parsing, pedestrian detection and instance segmentation, 2D/3D human pose estimation, and activity recognition.
\vspace{-2em}\end{IEEEbiography} 
\end{document}